\pdfoutput=1

\documentclass[11pt]{article}

\usepackage[final]{acl}

\usepackage{times}
\usepackage{latexsym}
\usepackage{enumitem}
\usepackage{amsmath}
\usepackage{multicol, multirow}
\usepackage{array}
\usepackage[table]{xcolor}
\usepackage{booktabs}
\usepackage{amssymb}

\usepackage[T1]{fontenc}

\usepackage[utf8]{inputenc}

\usepackage{microtype}

\usepackage{inconsolata}

\usepackage{graphicx}
\usepackage{authblk}  
\usepackage{todonotes}
\title{Target-Augmented Shared Fusion-based Multimodal Sarcasm \\Explanation Generation}

\author{\textbf{Palaash Goel$^1$\thanks{\textsuperscript{ } The first two authors contributed equally to this work and are jointly the first authors.}, Dushyant Singh Chauhan$^{2*}$, Md Shad Akhtar$^1$} \\
  $^1$ Indraprastha Institute of Information Technology Delhi, India \\
  $^2$ Indian Institute of Technology Patna, India \\
  {\tt \{palaash21547, shad.akhtar\}@iiitd.ac.in, 1821CS17@iitp.ac.in }}

\date{}

\newcommand{\model}{\texttt{TURBO}}
\newcommand{\dataset}{\texttt{MORE+}}

\begin{document}
\maketitle
\begin{abstract}

Sarcasm is a linguistic phenomenon that intends to ridicule a target (e.g., entity, event, or person) in an inherent way. Multimodal Sarcasm Explanation (MuSE) aims at revealing the intended irony in a sarcastic post using a natural language explanation. Though important, existing systems overlooked the significance of the target of sarcasm in generating explanations. In this paper, we propose a \textbf{T}arget-a\textbf{U}gmented sha\textbf{R}ed fusion-\textbf{B}ased sarcasm explanati\textbf{O}n model, aka. \model. We design a novel shared-fusion mechanism to leverage the inter-modality relationships between an image and its caption. \model\ assumes the target of the sarcasm and guides the multimodal shared fusion mechanism in learning intricacies of the intended irony for explanations. We evaluate our proposed \model\ model on the \dataset\ dataset. Comparison against multiple baselines and state-of-the-art models signifies the performance improvement of \model\ by an average margin of $+3.3\%$. Moreover, we explore LLMs in zero and one-shot settings for our task and observe that LLM-generated explanation, though remarkable, often fails to capture the critical nuances of the sarcasm. Furthermore, we supplement our study with extensive human evaluation on \model's generated explanations and find them out to be comparatively better than other systems.

\end{abstract}

\section{Introduction}\label{sec:intro}
Sarcasm is a form of communication usually involving statements meant to insult or mock some targets, such as a person, an entity, or an event. These statements point to one interpretation when taken literally but given the context within which they are uttered, mean something completely different. In other words, there is an incongruity between the explicit and implicit meanings of a sarcastic utterance. Resolving this incongruity is important for properly interpreting a sarcastic message. Existing research suggests significant dependence on cues from multiple sources to interpret sarcastic messages. These can include tone of voice, body language, common sense, etc. Furthermore, the task of identifying and understanding sarcasm is quite relevant in a multimodal scenario where each modality refers to a different source of sarcastic cues.

Recognizing this, \newcite{Desai_Chakraborty_Akhtar_2022} proposed the task of multimodal sarcasm explanation. They utilized multimodal social media posts, consisting of visual and textual information, and generated an explanation to reveal the intended irony in the post. 
Along with proposing the novel MORE dataset for this task, they also benchmarked this dataset using their proposed explanation model, ExMORE.
However, \newcite{jing-etal-2023-multi} recognized three limitations in ExMORE. They resolved these limitations by proposing their novel model, TEAM. The authors incorporated external world knowledge to aid sarcasm reasoning with the help of an innovative multi-source semantic graph. This allowed TEAM to perform exceptionally well, beating all baselines by significant margins. 
\begin{figure}[t]
  \includegraphics[width=\columnwidth]{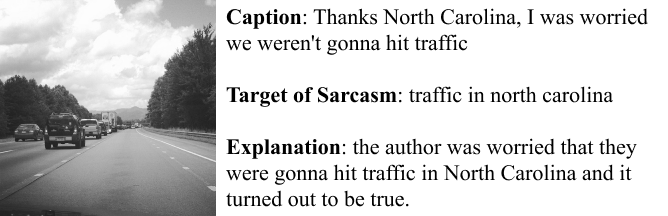}
  \caption{A sample in the \dataset\ dataset.}
  \label{fig:example-post}
\end{figure}

\paragraph{Motivation:} Despite its excellent performance, we identify two primary limitations in TEAM: \textbf{a)} \textit{ignoring vital visual cues}; and \textbf{b)} \textit{treating each node in the semantic knowledge relation as equivalent}. For the first case, TEAM converted the visual information into text-based metadata, which was subsequently used as a representation of all of the salient visual features. We argue that such an approach may lead to omitting vital visual features in the textual meta-data, representing the image. We hypothesize that both unimodal features, textual and visual, have their own significance in learning the multimodal semantics, leading to better explanations. For the second case, TEAM \cite{jing-etal-2023-multi} treated each extracted relation equivalent to each other by assigning them unit weights. However, we argue that some relations play relatively more significant roles than other relations in aiding sarcasm understanding.

We also identify the significance of the target of sarcasm in the explanations, as they are primarily intended to reveal the hidden irony towards the target. We observe that existing systems often fail to generate a relevant explanation when the target of sarcasm is highly implicit and needs a higher degree of cognition for comprehension. Therefore, we hypothesize that information about the target would lead to better comprehension by the model; and thus, improve the quality of explanation generation. Figure \ref{fig:example-post} depicts an instance of multimodal sarcasm explanation in the \dataset\ dataset. In the example, the user expresses the concerns towards the traffic in North Carolina (target of sarcasm) as conveyed by the explanation. We observe that the target of sarcasm is an inherent component of the expressed irony and must be exploited appropriately. 

Motivated by these limitations, we propose a novel method, \model\footnote{\textbf{T}arget-a\textbf{U}gmented sha\textbf{R}ed fusion-\textbf{B}ased sarcasm explanati\textbf{O}n}, to mitigate the challenges encountered by existing systems. \model\ comprises of three major components: \textbf{a)} \textit{knowledge-infusion for exploiting the external knowledge concepts considering their relevance}; \textbf{b)} \textit{a novel shared-fusion mechanism to obtain the multimodal fused representation}; and \textbf{c)} \textit{the target of sarcasm to guide the focused explanation generation}. To test our hypothesis, we extend the existing MORE dataset \cite{Desai_Chakraborty_Akhtar_2022} with manually annotated target label for the implied sarcasm. We call this extended dataset, \dataset. Our evaluation demonstrates the superiority of \model\ against the current state-of-the-art model, TEAM \cite{jing-etal-2023-multi}, and other baselines in both comparative analysis and human evaluation. We also explore the application of the Multimodal Large Language Models (MLLMs), such as, GPT-4o Mini, Llava-Mistral, and Llava-Llama-3, for the sarcasm explanation. We conclude the study with a qualitative error analysis and a human evaluation.



\paragraph{Contributions:} We summarize our contribution as follows: 

\begin{itemize}[leftmargin=*, noitemsep, nolistsep]
    \item \textbf{\dataset:} We extend the MORE dataset with target of sarcasm labels for 3,510 sarcastic posts. 
    \item \textbf{\model:} We propose a novel framework for multimodal sarcasm explanation. \model\ exploits the presence of target of sarcasm, external knowledge concepts, and a novel shared-fusion mechanism. 
    \item \textbf{Qualitative Analysis:} We perform extensive qualitative analysis in terms of error analysis and human evaluation.
\end{itemize}

\paragraph{Reproducibility:} Code and dataset are available at \url{https://github.com/flamenlp/TURBO}.

\section{Related Work}\label{sec:lit}
\begin{table*}[t!]
  \setlength{\tabcolsep}{1mm}
    \small
  \centering
      \begin{tabular}{lccccccc}
    \toprule
    \multirow{2}{*}{\bf Name} & \multirow{2}{*}{\bf \#Samples} & \multicolumn{2}{c}{\bf Caption}             & \multicolumn{2}{c}{\bf Explanation}        & \multicolumn{2}{c}{\bf Target of Sarcasm}  \\ \cmidrule{3-8} 
                          &                           & {Avg. length} & $|\nu|$    & {Avg. length} & $|\nu|$   & {Avg. length} & $|\nu|$  \\ \midrule
    \bf Train                 & 2,983                     & {19.75}      & 9,677  & {15.47}      & 5,972 & {4.17}       & 3776 \\ 
    \bf Validation                   & 175                       & {18.85}      & 1,230  & {15.39}      & 922   & {4.46}       & 452  \\ 
    \bf Test                  & 352                       & {19.43}      & 2,172  & {15.08}      & 1,527 & {4.57}       & 832  \\ 
    \bf Total                 & 3,510                     & {19.68}      & 10,865 & {15.43}      & 6,669 & {4.22}       & 4233 \\ \bottomrule
    \end{tabular}
  \caption{Statistical analysis of the \dataset\ dataset. $|\nu|$ denotes the size of the vocabulary.}
  \label{table:datset-stats}
\end{table*}
Sarcasm detection is a task that involves detecting whether sarcasm is present in a sample or not. The reason behind what makes a sample sarcastic is not in the scope of this task. In the beginning, researchers focused only on the textual modality for this task. For instance, \citet{Bouazizi-etal-sarcasm-detection-2016} and \citet{felbo-etal-2017-using} explored using hand-crafted features such as emojis, certain punctuation marks, and lexicons to detect whether an utterance is sarcastic or not. Neural network-based architectures have been used here as well (\citealp{tay-etal-2018-reasoning}; \citealp{babanejad-etal-2020-affective}). 

However, recognizing the multimodal nature of online content, \citet{schifanella-etal-2016} proposed the task of multimodal sarcasm detection. They proposed a system that uses Convolutional Neural Networks \citep{ma-etal-cnn-2015} to fuse textual and visual information and detect the presence of sarcasm. Having identified some limitations with this work, \citet{Qiao_Jing_Song_Chen_Zhu_Nie_2023}, \citet{kumar-etal-2022-become} and \citet{chakrabarty-etal-2020-r} explored ways to resolve them using Graph Convolutional Networks \citep{kipf2017semisupervised}. \citet{castro-etal-2019-towards} extended this task to conversational dialog systems and even proposed a new dataset, MUStARD, for the same. \citet{bedi-etal-code-mixed-2023} explored sarcasm detection in Hindi-English code-mixed conversations.

Researchers have also explored sarcasm analysis from a generative perspective, using only textual inputs in the beginning. For example, \citet{peled-reichart-2017-sarcasm} and \citet{dubey-2019-sarcasm-generation} explored the conversion of sarcastic texts into their non-sarcastic interpretations using machine translation techniques. On the contrary, \citet{mishra-etal-2019-modular} have worked on using sentences that express a negative sentiment to generate corresponding sarcastic text. 

\citet{Desai_Chakraborty_Akhtar_2022} proposed the task of multimodal sarcasm explanation (MuSE) and also proposed the MORE dataset for the same. Additionally, they proposed a model to benchmark MORE on this task, called ExMORE, which was built on a BART \citep{lewis-etal-2020-bart} backbone. 
However, ExMORE suffered from some limitations which were addressed by \citet{jing-etal-2023-multi} in their model, TEAM, which exceeded the performance of all existing baselines by large margins. Despite this, TEAM suffers from some limitations discussed in Section~\ref{sec:intro}. 
\section{Dataset}\label{sec:dataset}
We use the \dataset\ dataset, an extension of the MORE dataset \citep{Desai_Chakraborty_Akhtar_2022}, for conducting our experiments. The samples of this dataset are sarcastic posts from various online social media platforms such as Twitter, Instagram, and Tumblr. Each sample $i$ consists of an image ($V_i$) and a corresponding text caption ($C_i$). The sarcasm explanation for each sample has been manually annotated. 

We extend the MORE dataset by manually annotating the entity being targeted by sarcasm in each sample as ``target of sarcasm''. To ensure the quality of annotations, the following definitions and guidelines are followed.
\begin{itemize}[leftmargin=1em, noitemsep, nolistsep]
    \item The target of sarcasm is a short phrase that denotes an \textit{entity, phenomenon, concept, or fact being ridiculed with the use of sarcasm}.
    \item This phrase must not reveal the underlying sarcastic incongruity under any circumstances. 
    \item If two phrases describe the same target, the shorter and more straightforward phrase is preferable. 
    \item Any entity not being directly targeted by sarcasm must not be included. 
\end{itemize}
A detailed statistical analysis of the \dataset\ dataset is given in Table \ref{table:datset-stats}.
\section{Methodology}\label{sec:method}

\begin{figure*}[t]
  \includegraphics[width=\linewidth]{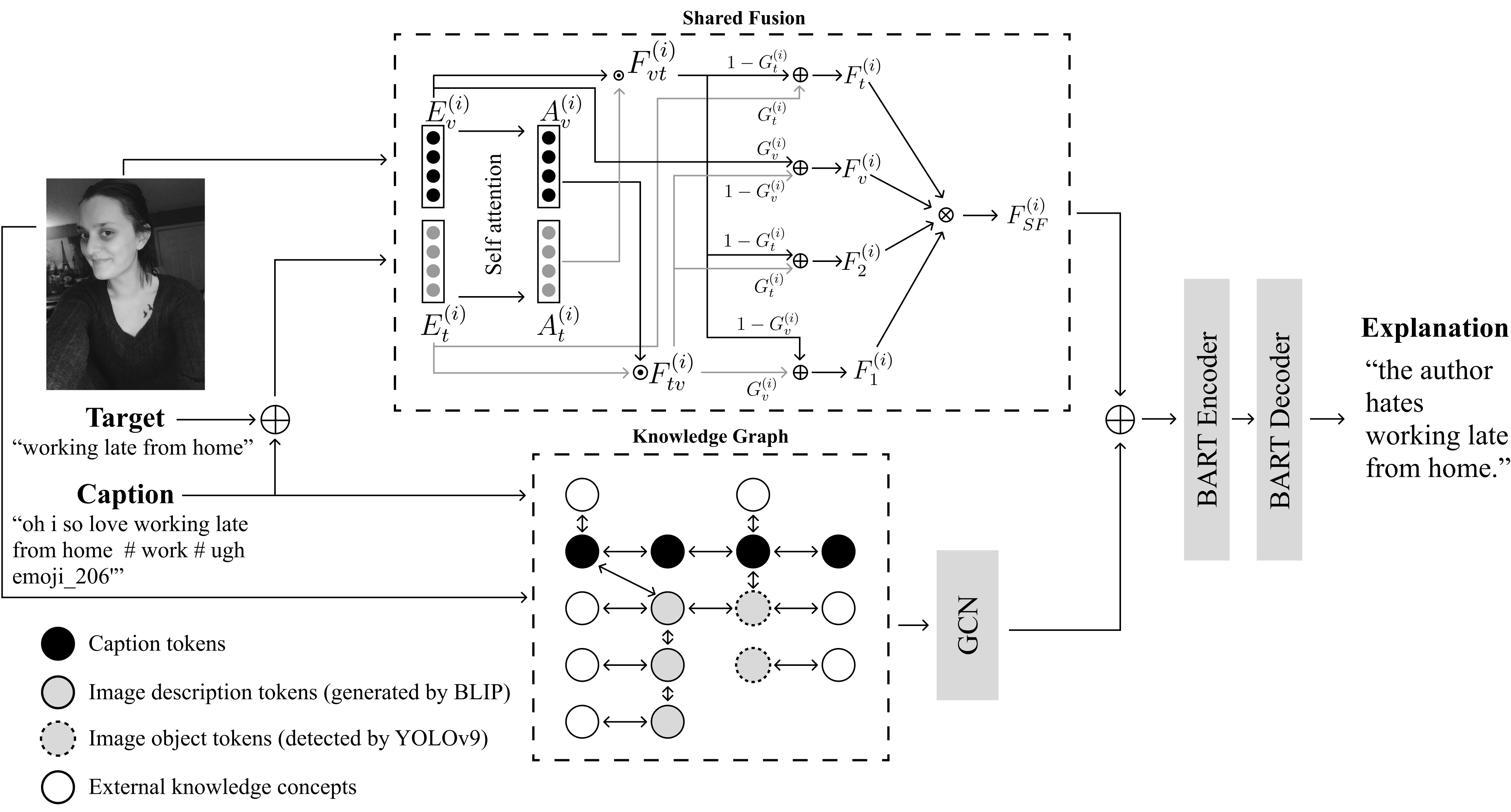}
  \caption{A schematic diagram of \model.}
  \label{fig:CASH}
\end{figure*}

In this section, we describe the proposed model and give a detailed account of its various components.

\subsection{Extraction of Visual Semantics} \label{sec:visual-semantics-extraction}
Considering the complex nature of sarcasm, the extraction of relevant visual information is quite important. 
In some samples, the sarcastic incongruity may be more relevant to a broad detail present in the image. However, in other samples, the sarcasm may become evident only after considering a small visual detail. Considering such diverse cases, we extract information from the visual modality at three different levels of granularity. 

\paragraph{Low-Level Detail:} This includes visual features that only provide an overview of an image. We capture such features by generating a single natural language description for each image. For this, we use the large variant of the BLIP model for image captioning\footnote{\url{https://huggingface.co/Salesforce/blip-image-captioning-large}} \citep{li2022blip}. Formally,
\begin{equation}
    \label{eq: coarse-grained-semantics}
    \begin{aligned}
        BLIP(V_i) = D_i = \{d_1^{(i)}, d_2^{(i)}, \dots d_{N_{D_i}}^{(i)}\}\\
    \end{aligned}
\end{equation}
where $D_i$ is the generated description of $V_i$. Also, $d_j^{(i)}$ is the $j^{th}$ token present in $D_i$.
        
\paragraph{Medium-Level Detail:} Though low-level details provide helpful information to the model and facilitate its understanding of the underlying semantic incongruity, such generated descriptions may not capture the entire semantic meaning present in an image. Consequently, it leads us to capture information about individual entities appearing in an image as medium-level details. Identifying such entities in an image helps the model analyze a more detailed view of it. These entities are extracted by performing object detection on each image using the YOLOv9 object detection model \cite{wang2024yolov9}. We keep the top $K$ objects with the highest confidence to ensure that only the most relevant object-level semantics are retained.
\begin{equation}
    \label{eq: medium-grained-semantics}
    \begin{aligned}
        YOLOv9(V_i) = O_i = \{o_1^{(i)}, o_2^{(i)}, \dots o_K^{(i)}\}\\
    \end{aligned}
\end{equation}
where $o_j^{(i)}$ is a text label describing the $j^{th}$ object extracted from $V_i$ using YOLOv9. 

\paragraph{High-Level Detail:} Finally, we utilize semantic-rich representations of visual features. These are extracted in the form of embedding vectors from a pre-trained vision transformer \cite{dosovitskiy2021image}.
\begin{equation}
    \label{eq: fine-grained-semantics}
    \begin{aligned}
        ViT(V_i) = E_v^{(i)} =
        \begin{bmatrix}
            \textbf{v}_1^{(i)} & \textbf{v}_2^{(i)} & \dots & \textbf{v}_m^{(i)}
        \end{bmatrix}^T
    \end{aligned}
\end{equation}
where $E_v^{(i)} \in \mathbb{R}^{m \times D_f}$ is the feature matrix of $V_i$ containing ``$m$'' $D_f$-dimensional vectors.
\subsection{External Knowledge Retrieval}
ConceptNet\footnote{\url{https://conceptnet.io/}} \cite{speer2017conceptnet} is a knowledge graph that structures general human knowledge as a directed and weighted graph. We use it to extract relevant external knowledge concepts for our model. 

For the $i^{th}$ sample, we retrieve external knowledge concepts related to caption tokens in $C_i$, $D_i$ (\S Equation~\ref{eq: coarse-grained-semantics}) and $O_i$ (\S Equation~\ref{eq: medium-grained-semantics}). Each token is queried through ConceptNet and we utilize its one-hop neighboring knowledge concept along with its relevance score. 

Note that frequent words\footnote{NLTK stopwords \cite{nltk}} such as `the', `and', `is', etc. are not queried. Formally, 
\begin{equation}
    \label{eq: external-knowledge-retrieval}
    \begin{aligned}
        EK(t_i) & = ConceptNet(t_i)_1 = (T_1^{(i)}, r_1^{(i)})
    \end{aligned}
\end{equation}
where $t_i \notin StopWords$ is a queried token and $T_1^{(i)}$ is its one-hop neighbouring external knowledge concept with a relevance score of $r_1^{(i)}$.

\subsection{Knowledge Enrichment}
\label{sec:knowledge-enrichment}
For each sample, we first 
perform string concatenation on $C_i$, $D_i$, $O_i$, and their related knowledge concepts. The resultant string is called $T_{knowledge}$. This is a knowledge-enriched sequence of tokens that provides the model with more information than what is present in just $C_i$. Formally,
\begin{equation}
    \label{eq: t-knowledge}
    \begin{aligned}
        T_{knowledge}^{(i)} = C_i + CC_i + D_i + DC_i + O_i + OC_i
    \end{aligned}
\end{equation}
where $CC_i$, $DC_i$ and $OC_i$ are the knowledge concepts corresponding to the tokens of $C_i$, $D_i$ and $O_i$, respectively.

We put a constraint on the ordering of the individual tokens in $CC_i$, $DC_i$, and $OC_i$ to avoid random permutations. If $c_j^{(i)}, c_k^{(i)} \in C_i$ and their respective knowledge concepts are $cc_p^{(i)}, cc_q^{(i)} \in CC_i$. Then,
\begin{equation}
    \label{eq: knowledge-tokens-constraints}
    \begin{aligned}
        j < k \iff p < q
    \end{aligned}
\end{equation}

Similar constraints are placed on $DC_i$ and $OC_i$.

\subsection{Construction of Knowledge Graph}
\label{sec:graph-construction}
$T_{knowledge}$ simply provides the model with additional information in the form of a sequence of tokens. However, the relationships among these tokens are often non-sequential and would be much better represented by a non-linear data structure such as a graph. 

Therefore, we build an undirected, weighted graph $G$ of these tokens. Not only is this a more appropriate representation of the various inter-token relationships, it also facilitates our model's understanding of them. We adopt the following strategy to construct it. 

\begin{enumerate}
    \item We construct an edge of unit weight between every pair of consecutive tokens in $C_i$. 
    \begin{equation}
      \label{eq:connect-input-caption}
      \begin{aligned}
        \forall j: e(c_j^{(i)}, c_{j+1}^{(i)}) = 1 
      \end{aligned}
    \end{equation}
    
    where $c_j^{(i)}$ is the $j^{th}$ token in $C_i$ and $e(a,b)$ denotes the weight of the edge linking nodes $a$ and $b$.
    \item Each $c_j^{(i)}$ is then linked with its corresponding external knowledge concept ($cc_j^{(i)}$) with an edge of a weight equal to the corresponding relevance score ($r_j^{(i)}$). This helps in capturing the strength of the relationship between the nodes.   
    \begin{equation}
      \label{eq:caption-concepts}
      \begin{aligned}
        \forall j: e(c_j^{(i)}, cc_j^{(i)}) = r_j^{(i)}
      \end{aligned}
    \end{equation}
    
    \item A similar strategy is adopted for $D_i$ and $O_i$.
    \begin{equation}
      \label{eq:image-caption-connections}
      \begin{aligned}
        \forall j: e(d_j^{(i)}, d_{j+1}^{(i)}) & = 1 \\
        \forall j: e(d_j^{(i)}, dc_j^{(i)}) &  = r_j^{(i)} \\
        \forall j: e(o_j^{(i)}, oc_j^{(i)}) & = r_j^{(i)}
      \end{aligned}
    \end{equation}
    
 \end{enumerate}   

\noindent Note that the object tokens do not follow a syntactic ordering; therefore, we do not construct edges between consecutive object tokens. 
    
    



\subsection{Incorporation of Target of Sarcasm}
\label{sec:incorporation-of-target}
The target of sarcasm is incorporated in the model by concatenating it with $T_{knowledge}$ as follows:
\begin{equation}
\label{eq:target-incorporation}
T_{concat}^{(i)} = T_{knowledge}^{(i)} + \verb|</s>| + TS_i
\end{equation}

\noindent where $TS_i$ is the target of sarcasm for the $i^{th}$ sample and \verb|</s>| is the BART separator token. $T_{concat}$ is the final sequence of tokens that is fed to the model as input along with the sample image. We use BART to extract contextual embeddings, $E_t$, for $T_{concat}$ where $E_t \in \mathbb{R}^{N \times D_f}$.

\subsection{Sarcasm Reasoning}
\label{sec:graph-convolution}
In order to facilitate sarcasm reasoning and extract the salient features from our knowledge graph, we follow \newcite{jing-etal-2023-multi} and use a Graph Convolution Network (GCN) \cite{kipf2017semisupervised}. We incorporate $L$ GCN layers in our model where the output of layer $l$ is computed as follows. 
\begin{equation}
  \label{eq:gcn}
  \begin{aligned}
    H_l^{(i)} = f(\hat{D_i}^{-\frac{1}{2}}\hat{A_i}\hat{D_i}^{-\frac{1}{2}}H_{l-1}^{(i)}W_l)
  \end{aligned}
\end{equation}

\noindent where, for the $i^{th}$ sample, $H_l^{(i)}$ is the output of layer $l$ and $\hat{A_i}$ is the adjacency matrix corresponding to the knowledge graph. $\hat{D_i}$ is a diagonal matrix that stores the degree of each node in it. $W_l$ is a learnable weight matrix and $f$ is a non-linear activation function. Additionally, $H_0^{(i)} = E_t^{(i)}$.

$H_L^{(i)}$ is the final output of the GCN and provides a representation of the salient semantic features derived from the nodes of the knowledge graph. 

\subsection{Shared Fusion}
In order to facilitate the sharing of information between and within modalities, we propose a novel shared fusion mechanism. We first perform self-attention \citep{attention-is-all-you-need} on the text and image embeddings which allows the model to capture important intra-modal semantic relationships.  
\begin{equation}
  \label{eq:Self Attention}
  \begin{aligned}
  A_v^{(i)} & = Softmax(\frac{Q_v^{(i)} K_v^{(i)^T}}{d_k})V_v^{(i)} \\
  A_t^{(i)} & = Softmax(\frac{Q_t^{(i)} K_t^{(i)^T}}{d_k})V_t^{(i)}
  \end{aligned}
\end{equation}
\noindent where $K_v$, $Q_v$ and $V_v$ are the key, query, and value matrices used while computing self-attention on the visual embeddings. Similarly, $K_t$, $Q_t$, and $V_t$ are used to compute self-attention on text embeddings. 

We use $A_v^{(i)}$ to amplify relevant features represented in $E_t^{(i)}$ and vice versa. This facilitates a cross-modal flow of information and aids multimodal learning as follows:
\begin{equation}
  \label{eq:Fit-Fti}
  \begin{aligned}
  F_{vt}^{(i)} = A_t^{(i)} \odot E_v^{(i)} ; \quad F_{tv}^{(i)} = A_v^{(i)} \odot E_t^{(i)}
  \end{aligned}
\end{equation}

\noindent where $\odot$ denotes element-wise multiplication. 

While $F_{vt}$ and $F_{tv}$ capture cross-modal relationships, we need a mechanism that can dynamically control both the cross-modal and intra-modal flow of information. This is because, for different samples, 
each modality may contribute differently towards the sarcastic incongruity. In some samples, the image may contribute more towards the sarcastic incongruity while in others, the text caption may be a bigger factor. Therefore, we need a mechanism that ensures that the model does not overly rely on either modality. 

We implement such a mechanism in the form of a gated fusion mechanism. Here, $F_{vt}$ \& $F_{tv}$ serve as the cross-modal feature representations and $E_v$ \& $E_t$ serve as the unimodal feature representations. The gating weights for this mechanism control the flow of information between these sources of information and are computed as follows:
\begin{equation}
    \label{eq: gated-fusion-weights}
    \begin{aligned}
        G_v^{(i)} & = \sigma(W_v^{(i)}E_v^{(i)} + b_v^{(i)}) \\  G_t^{(i)} & = \sigma(W_t^{(i)}E_t^{(i)} + b_t^{(i)})
    \end{aligned} \nonumber
\end{equation}
where $W_v^{(i)}$ and $W_t^{(i)}$ are learnable weight matrices and $b_v^{(i)}$ and $b_t^{(i)}$ are their corresponding biases. $\sigma$ refers to the sigmoid function. 

Using $G_v^{(i)}$ and $G_t^{(i)}$, we compute the final shared fusion matrix as a weighted sum of four individual gated fusions. These are given as follows.
\paragraph{Fusing Two Multimodal Representations.} The semantic relationships and features captured by $F_{tv}$ will be different from those captured by $F_{vt}$. Thus, we use $G_v$ and $G_t$ to separately weigh the semantic information present in them. This allows us to capture the salient features in both multimodal representations.
\begin{equation}
    \label{eq: F1-F2}
    \begin{aligned}
        F_1^{(i)} = (G_v^{(i)} \odot F_{tv}^{(i)}) + [(1-G_v^{(i)}) \odot F_{vt}^{(i)}] \\
        F_2^{(i)} = (G_t^{(i)} \odot F_{tv}^{(i)}) + [(1-G_t^{(i)}) \odot F_{vt}^{(i)}]
    \end{aligned}
\end{equation}

\paragraph{Fusing Multimodal and Unimodal Representations.} It is also possible that by bringing the salient features of both modalities together, the sarcastic incongruity may become less evident than if just one of the modalities was considered. Thus, we compare the semantic information present in an unimodal representation with that present in a multimodal representation. 

Specifically, we first use $G_v$ to weigh the relevant features present in just the visual modality ($E_v$) as well as those present in $F_{tv}$.  
\begin{equation}
    \label{eq: Fv}
    \begin{aligned}
        F_v^{(i)} = (G_v^{(i)} \odot E_v^{(i)}) + [(1-G_v^{(i)}) \odot F_{tv}^{(i)}]
    \end{aligned}
\end{equation}

We then compute $F_t$ by performing a similar computation as above, with the textual modality. 
\begin{equation}
    \label{eq: Ft}
    \begin{aligned}
        F_t^{(i)} = (G_t^{(i)} \odot E_t^{(i)}) + [(1-G_t^{(i)}) \odot F_{vt}^{(i)}]
    \end{aligned}
\end{equation}

We combine $F_1$, $F_2$, $F_v$, and $F_t$ to allow the model to weigh the salient features captured in each of these matrices. This leads to a multimodal representation containing only the most semantically relevant information with respect to the underlying sarcasm. 
\begin{equation}
    \label{eq: sf-out}
    \begin{aligned}
        F_{SF}^{(i)} = \alpha_1F_1^{(i)} + \alpha_2F_2^{(i)} + \beta_1F_v^{(i)} + \beta_2F_t^{(i)}
    \end{aligned}
\end{equation}

where $\alpha_1$, $\alpha_2$, $\beta_1$, and $\beta_2$ are learnable parameters allowing the model to dynamically weigh the information present in the four individual matrices. 

\subsection{Sarcasm Explanation Generation}
In order to generate the sarcasm explanations, we sum $H_L$ and $F_{SF}$.
\begin{equation}
    \label{eq: encoder-inp}
    \begin{aligned}
        Z_i = H_L^{(i)} + F_{SF}^{(i)}
    \end{aligned}
\end{equation}

We pass $Z$ to BART to fine-tune it on the \dataset\ dataset for generating sarcasm explanations. The self-attention layer in its encoder captures various semantic relationships and dependencies while the decoder generates the explanations in an auto-regressive manner, that is, by taking the previously generated words into account when predicting the next word in a sequence. 



\section{Experiments, Results, and Analyses}\label{sec:experiments}

\begin{table*}
\setlength{\tabcolsep}{1mm}
  \centering
  \resizebox{\textwidth}{!}{
  \begin{tabular}{lcccccccccccc}
    \toprule
     \multirow{2}{*}{\bf Models} & \multicolumn{4}{c}{\bf BLEU} & \multicolumn{3}{c}{\bf ROUGE} &  \multirow{2}{*}{\bf METEOR} & \multicolumn{3}{c}{\bf BERTScore} & \multirow{2}{*}{\bf SentBERT} \\ \cmidrule{2-5} \cmidrule{6-8} \cmidrule{10-12}
     
    & \bf B1 & \bf B2 & \bf B3 & \bf B4 & \bf RL & \bf R1 & \bf R2 & & \bf Precision & \bf Recall & \bf F1 & \\
    \midrule
    \textbf{PGN} & 17.54 & 6.31 & 2.33 & 1.67 & 16.00 & 17.35 & 6.90 & 15.06 & 84.80 & 85.10 & 84.90 & 49.42\\
    
    \textbf{Transformer} & 11.44 & 4.79 & 1.68 & 0.73 & 15.90 & 17.78 & 5.83 & 9.74 & 83.40 & 84.90 & 84.10 & 52.55\\
    \textbf{MFFG-RNN} & 14.16 & 6.10 & 2.31 & 1.12 & 16.21 & 17.47 & 5.53 & 12.31 & 81.50 & 84.00 & 82.70 & 44.65\\
    \textbf{MFFG-Transf} & 13.55 & 4.95 & 2.00 & 0.76 & 15.14 & 16.84 & 4.30 & 10.97 & 81.10 & 83.80 & 82.40 & 41.58\\
    \textbf{M-Transf} & 14.37 & 6.48 & 2.94 & 1.57 & 18.77 & 20.99 & 6.98 & 12.84 & 86.30 & 86.20 & 86.20 & 53.85\\
    \textbf{ExMore} & 19.26 & 11.21 & 6.56 & 4.26 & 25.23 & 27.55 & 12.49 & 19.16 & 88.30 & 87.50 & 87.90 & 59.12\\ \midrule
    \textbf{GPT-4o-Mini-Zeroshot} & 17.51 & 7.71 & 3.98 & 2.04 & 20.68 & 21.77 & 5.20 & 26.45 & 85.47 & 87.41 & 86.42 & 56.23 \\
    \textbf{LLaVa-Mistral-Zeroshot} & 17.25 & 8.50 & 5.16 & 3.38 & 22.32 & 23.03 & 6.51 & 27.47 & 85.60 & 87.80 & 86.68 & 58.97 \\
    \textbf{LLaVa-Llama3-Zeroshot} & 21.08 & 11.08 & 7.00 & 4.81 & 25.30 & 25.76 & 8.21 & 28.33 & 86.38 & 87.82 & 87.08 & 60.06 \\ \midrule    
    \textbf{GPT-4o-Mini-Oneshot} & 19.07 & 8.45 & 4.33 & 2.23 & 22.01 & 22.68 & 5.68 & 25.89 & 86.01 & 87.61 & 86.79 & 57.30 \\
    \textbf{LLaVa-Mistral-Oneshot} & 20.29 & 10.04 & 6.03 & 3.93 & 24.57 & 24.47 & 6.75 & 26.24 & 86.46 & 87.85 & 87.13 & 59.76 \\
    \textbf{LLaVa-Llama3-Oneshot} & 24.50 & 13.12 & 8.31 & 5.58 & 27.68 & 27.77 & 8.80 & 27.60 & 87.07 & 87.85 & 87.44 & 61.07 \\ \midrule
    \textbf{TEAM [SOTA]} & 55.32 & 45.12 & 38.27 & 33.16 & 50.58 & 51.72 & 34.96 & 50.95 & 91.80 & 91.60 & 91.70 & 72.92 \\
    \midrule
    
    \rowcolor{red!20} \textbf{TURBO} & \textbf{57.09} & \textbf{46.93} & \textbf{40.28} & \textbf{35.26} & \textbf{53.12} & \textbf{55.06} & \textbf{38.16} & \textbf{55.17} & \textbf{92.00} & \textbf{91.77} & \textbf{91.86} & \textbf{75.75}\\     \midrule
    \textbf{TURBO $+$ TS Concepts} & 55.65 & 45.38 & 38.55 & 33.36 & 51.64 & 53.63 & 36.50 & 53.80 & 91.72 & 91.60 & 91.64 & 75.15\\

    \textbf{TURBO} $-$ \textbf{SF} $-$ \textbf{TS} & 51.54 & 39.85 & 32.30 & 26.79 & 47.62 & 49.43 & 30.66 & 48.86 & 91.17 &	90.67 &	90.90 & 72.25\\
    \textbf{TURBO} $-$ \textbf{KG} $-$ \textbf{TS} & 54.19 & 44.17 & 37.73 & 32.93 & 49.80 & 50.72 & 34.42 & 51.08 & 91.42 & 91.22 & 91.29 & 73.20\\
    \textbf{TURBO} $-$ \textbf{TS} & 55.37 & 45.09 & 38.45 & 33.53 & 50.98 & 52.25 & 35.41 & 52.08 & 91.55 &	91.51 &	91.51 & 73.62\\
    \textbf{TURBO} $-$ \textbf{KG} & 55.56 & 44.80 & 37.71 & 32.58 & 51.42 & 53.46 & 35.46 & 53.22 & 91.76 &	91.65 &	91.68 & 75.16\\
    \textbf{TURBO} $-$ \textbf{SF} & 54.72 & 42.95 & 35.26 & 29.65 & 50.69 & 52.38 & 32.97 & 52.16 & 91.67 & 91.35 & 91.50 & 74.62\\  \bottomrule
  \end{tabular}}
  \caption{Results of a comparative analysis of our proposed model with multiple state-of-the-art baselines and an ablation study. These analyses were conducted on the \dataset\ dataset. The best results are in boldface.}
  \label{comparative-analysis}
\end{table*}

We built our model on top of the base variant of the BART model. The feature dimension ($D_f$) of this variant of BART is 768. For the textual modality, the total number of tokens in each sample ($N$) is set to 256 by utilizing padding and truncation operations wherever necessary. The pre-trained visual feature embeddings extracted from vision transformer (\S \ref{sec:visual-semantics-extraction}) are of dimensions $50 \times 768$ which are projected to a $256 \times 768$ embedding space using a learnable linear layer. Additionally, the maximum number of objects that can be extracted from an image, $K$, is set to 36.

We use a learning rate of $10^{-3}$ for the GCN layers and that of $10^{-4}$ for BART. We employ AdamW \cite{loshchilov-etal-adamw} as the optimizer and train our model for 20 epochs using the standard cross entropy loss on a batch size of 16. We train our model on a system running an Ubuntu operating system with a Tesla V100-PCIE-32GB GPU. The model requires approximately 9GB of RAM for the same. 


\subsection{Comparative Analysis}
\label{sec:comparative-analysis}

In line with the existing systems \cite{Desai_Chakraborty_Akhtar_2022,jiang-2023-team}, we employ BLEU, ROUGE, METEOR, BERTScore, and SentBERT for evaluation.
We compare the performance of \model\ with various unimodal and multimodal baselines. 
\begin{itemize}[leftmargin=*, noitemsep, nolistsep]
\item{
\textbf{Existing Baselines}. We employ all of the baselines employed by \citet{Desai_Chakraborty_Akhtar_2022}. 
These are: 
\textbf{a)} \textit{Pointer Generator Network (PGN)} \citep{see-etal-2017-get};
\textbf{b)} \textit{Transformer} \citep{attention-is-all-you-need};
\textbf{c)} \textit{MFFG-RNN and MFFG-Transf} \citep{liu-etal-2020-multistage};
\textbf{d)} \textit{M-Transf} \citep{yao-wan-2020-multimodal}; and
\textbf{e)} \textit{ExMore} \citep{Desai_Chakraborty_Akhtar_2022}.
}
\item{\textbf{TEAM} \citep{jing-etal-2023-multi}. This model refers to the current state-of-the-art in this task.}   
\end{itemize}
Notably, we also compare our model with multiple state-of-the-art multimodal large-language models (MLLMs) in zero-shot and one-shot settings, such as, \textbf{GPT-4o Mini} \cite{openai2024gpt4technicalreport}, \textbf{LLaVa-Mistral} (\citealp{liu2024improved}; \citealp{jiang2023mistral}), and \textbf{LLaVa-Llama3} (\citealp{liu2024improved}; \citealp{dubey2024llama3herdmodels}). The complexity and size of these LLMs, combined with the vast amount of data that they are trained on, allow them to perform exceptionally well on a diverse set of tasks. Thus, by including this comparison, we aim to demonstrate the effectiveness of \model\ and its superiority at performing this task even when compared to extremely powerful and versatile LLMs.  




\paragraph{Existing Baselines:} We conduct our experiments on the \dataset\ dataset and report our results in Table \ref{comparative-analysis}. We observe a significant disparity in the performance of our model when compared to non-LLM baselines other than TEAM. Moreover, compared to TEAM, we observe that \model\ consistently reports better scores on each evaluation metric -- \model\ outperforms TEAM on B1, B2, B3 and B4 by $+1.77\%$, $+1.81\%$, $+2.01\%$, and $+2.10\%$, respectively, with an average margin of $+1.92\%$. Furthermore, \model\ yields better ROUGE [RL ($+2.54\%$), R1 ($+3.34\%$), R2 ($+3.20\%$)] and METEOR [$+4.22\%$] scores of the two models with an average margin of $+3.33\%$. 
We also compute BERTscore and SentBERT to evaluate the semantic context in the generated explanations. We observe that our model surpasses TEAM by margins of $+0.2\%$, $+0.17\%$, and $0.16\%$ on the mean precision, recall, and F1 components of BERTscore (average margin of $+0.18\%$) and by a margin of $+2.83\%$ on SentBERT.

\paragraph{Multimodal LLMs:} For completeness, we also evaluate the MLLMs' explanations on the aforementioned evaluation metrics. As evident in Table \ref{comparative-analysis}, \model\ outperforms MLLMs on all evaluation metrics; however, we acknowledge that these metrics are not appropriate for assessing the quality and semantic richness of the generated explanations. Therefore, we also perform a human evaluation to better gauge their quality (\S \ref{sec:human-eval}).



\subsection{Ablation Study}
\label{sec:ablation}

We conduct an ablation study on various components of \model\ as \model\ $-$ X, where X corresponds to the Shared Fusion module (\textbf{SF}), the Knowledge Graph (\textbf{KG}), or the Target of Sarcasm (\textbf{TS}). Note that for each variant that does not contain the target of sarcasm, we simply avoid concatenating it with the other text tokens (\S \ref{sec:incorporation-of-target}). The results of the ablation study are given in Table \ref{comparative-analysis}. From the results of the ablation study, we make the following observations.

First, the removal of any component of the model leads to a decrease in its generative performance. This proves that each component of the model contributes significantly to the model's performance. Additionally, we observe that ``\model\ $-$ KG'' performs better than ``\model\ $-$ KG $-$ TS'' and that ``\model\ $-$ SF'' exceeds the performance of ``\model\ $-$ SF $-$ TS''. This demonstrates that the target of sarcasm is useful in facilitating the understanding of the underlying sarcastic incongruity, thus proving our hypothesis as given in Section \ref{sec:intro}.

Second, we compare \model\ with its ``\model\ + TS Concepts'' variant. As opposed to \model, this variant also utilizes external knowledge concepts for the target of sarcasm and helps us observe the effect of the same on the model's performance. For this variant, Equation \ref{eq:target-incorporation} is revised as follows to incorporate the knowledge concepts extracted for the target of sarcasm.
\begin{equation}
\label{eq:TS-concepts-incorporation}
T_{concat}^{(i)} = T_{knowledge}^{(i)} + \verb|</s>| + TS_i + TSC_i
\end{equation}
where $TSC_i$ refers to the external knowledge concepts for the target of sarcasm of the $i^{th}$ sample. Note that while the \model\ + TS Concepts variant was unable to perform as well as \model, it was still able to outperform TEAM on almost all metrics, which further proves the effectiveness of \model's architecture.

\subsection{Human Evaluation}
\label{sec:human-eval}
\begin{table}
\setlength{\tabcolsep}{1mm}
\small
  \centering
  \begin{tabular}{lcccc}
    \toprule

    \multirow{2}{*}{\textbf{Models}} & \multirow{2}{*}{\textbf{Fluency}} &  \bf Sem. & \multirow{2}{*}{\textbf{Negativity}} & \bf Target \\ 
     &  & \bf Acc. &  & \bf Presence \\

     \midrule
    \textbf{TURBO} & 4.18 & 3.80 & 3.82 & 4.00 \\ \midrule
    \textbf{TEAM} & 3.96 & 3.38 & 3.48 & 3.47\\ \midrule
    \textbf{GPT-4o-Mini} & 4.23 & 4.46 & 4.33 & 4.33\\
    \textbf{LLaVa-Mistral} & 4.26 & 4.09 & 4.22 & 4.03\\
    \textbf{LLaVa-Llama3} & 4.25 & 3.69 & 3.68 & 3.72\\
    \bottomrule   
  \end{tabular}
  \caption{Human evaluation on \model\ and other baselines. All LLMs are employed in a one-shot setting.}
  \label{table:human-eval}
\end{table}

\begin{figure*}[t]
  \includegraphics[width=\linewidth]{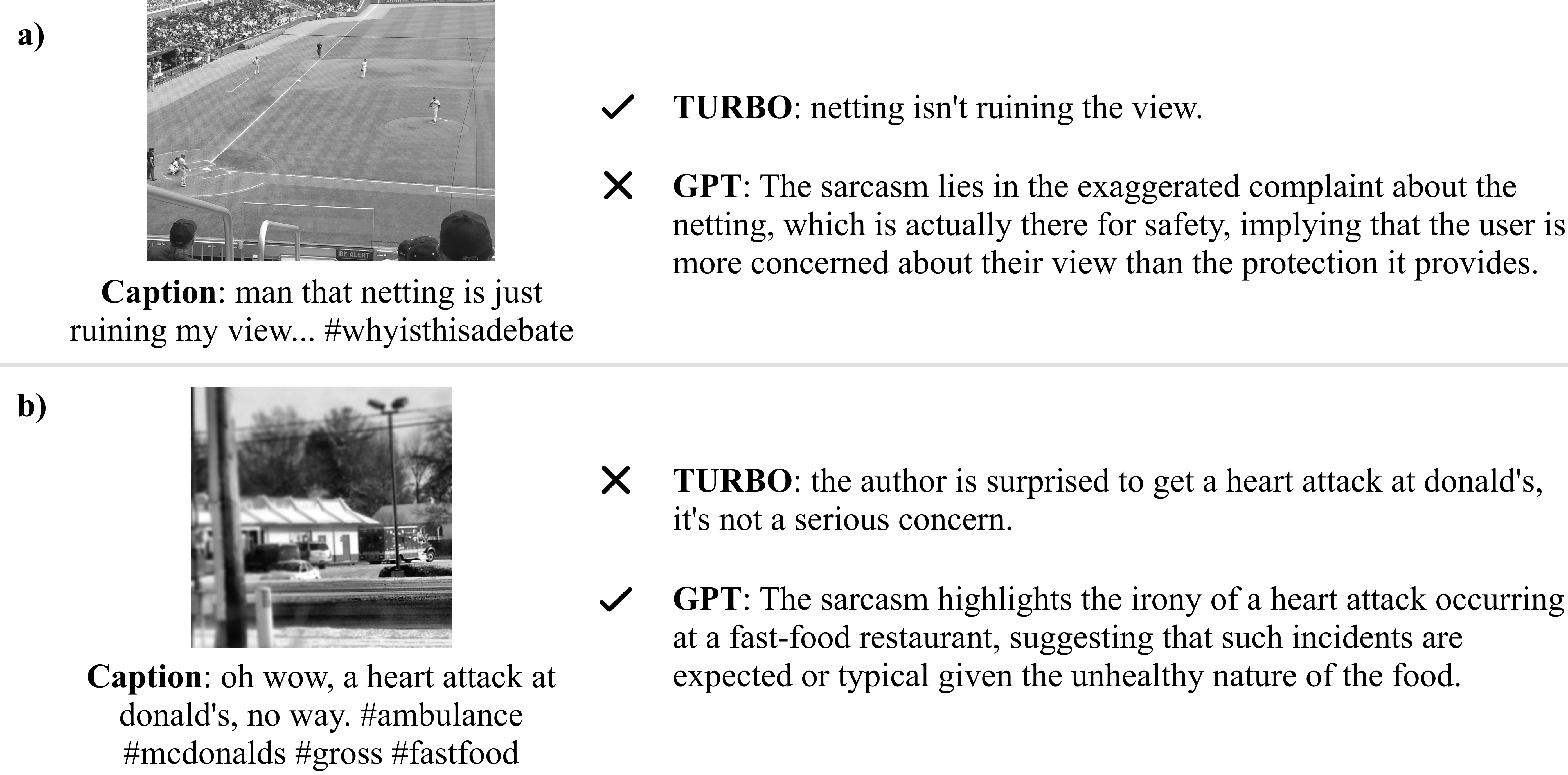}
  \caption{Two samples where: \textbf{a)} \model\ correctly explained the sarcasm and GPT-4o mini missed it; and \textbf{b)} \model\ generated an inaccurate explanation but GPT-4o mini explained the sarcasm correctly.}
  \label{fig:human-eval-qualitative-samples}
\end{figure*}

We conduct a human evaluation to assess the quality of generated explanations more comprehensively. We perform this evaluation on 20 diverse samples from the test set. These samples are assessed by 20 evaluators. For each sample, we provide the evaluator with five explanations, each generated by a different model. We ask the evaluator to judge each explanation on the following four metrics:
\begin{itemize}[leftmargin=*, noitemsep, nolistsep]
    \item{\textbf{Fluency:} This metric assesses \textit{how easy it is to understand the semantic meaning of a given explanation}. 
    }
    \item{\textbf{Semantic Accuracy:} This metric is used to determine \textit{how well a generation captures the intended meaning of a sarcastic post}.} 
    \item{\textbf{Negative Connotation:} Sarcasm expresses negative sentiment by default; hence, it must be captured in the explanation as well. We employ this metric to understand \textit{how appropriately the negative connotation is captured in the explanation}.}
    \item{ \textbf{Presence of Target:} This metric evaluates the presence of the target of sarcasm in the generated explanation. 
    }
\end{itemize}
The results for this evaluation are given in Table \ref{table:human-eval} with all metrics rated on a 5-point Likert scale. \model\ outperforms TEAM by 4.40\%, 8.40\%, 6.80\%, and 10.60\% on the above four metrics respectively. It even outperforms LLaVa-Llama3 on every metric except fluency, on which it is only 1.40\% worse. Additionally, we recognize that it performs worse than LLaVa-Mistral and GPT-4o-Mini by an average of 4.00\% and \textasciitilde 7.75\% respectively. However, we note that \model\ has 234 million parameters as compared to GPT-4o-Mini, LLaVa-Mistral, and LLaVa-Llama3 which have 7-8 billion parameters each. Therefore, with an approximately 3000\% reduction in the number of model parameters, \model\ demonstrates, at most, a 7.75\% dip in performance when compared to various state-of-the-art MLLMs. 


Figure \ref{fig:human-eval-qualitative-samples} depicts two samples and the explanations generated by \model\ and GPT-4o mini. In the first sample, the user sarcastically points out that the safety netting at a baseball game does not obstruct one's view and that people are needlessly complaining about the same. While \model\ is able to understand the sarcasm and correctly points out that the netting does not ruin one's view of the game, GPT-4o mini instead incorrectly identifies the user as someone who is concerned more about their view rather than their protection.

In the second sample, the author sarcastically points out how unhealthy fast food is by feigning surprise that someone got a heart attack at McDonald's. Here, \model\ mistakenly concludes that the author is surprised about someone getting a heart attack at McDonald's but does not think it is a serious concern. GPT-4o mini, however, is able to correctly understand the underlying sarcasm and generates an accurate explanation.

\subsection{Error Analysis} \label{sec:error-analysis}
Despite the excellent performance of \model, we recognize that it can exhibit different types of errors, which can lead to inaccurate sarcasm explanations. 
Therefore, we conduct a thorough error analysis and identify the following as the primary causes of erroneous behavior in \model's performance:
\textbf{a)} \textit{extraction of irrelevant external knowledge concepts} (c.f. Figure \ref{fig:error-analysis}); 
\textbf{b)} \textit{missing external knowledge concepts} (c.f. Figure \ref{fig:error-analysis-no-knowledge});
\textbf{c)} \textit{insufficient OCR features} (c.f. Figure \ref{fig:error-analysis-OCR});
\textbf{d)} \textit{irrelevant image description} (c.f. Figure \ref{fig:error-analysis-irrelevant-desc}). Appendix \ref{sec:appendix-error-analysis} provides a detailed account of the errors exhibited by \model.
\section{Conclusion} \label{sec:con}
In this research, we proposed a novel target-augmented shared fusion-based multimodal sarcasm explanation model, \model. Along with solving the key limitations that we were able to identify in the current state-of-the-art, TEAM, we introduced a novel method of fusing unimodal information into an effective multimodal feature representation. We also manually augmented the MORE dataset by annotating the target of sarcasm. Thorough comparative analyses allowed us to demonstrate the effectiveness of \model\ when compared to existing baselines, including various state-of-the-art MLLMs. Additionally, in order to comprehensively evaluate the generations of the MLLMs and those of \model, we conducted a human evaluation. Even though \model\ did not surpass the scores of two out of the three MLLMs in this evaluation, its results were quite comparable to theirs which is remarkable considering its much smaller size and lesser complexity.  
\section{Limitations} \label{sec:limitations}
Despite beating all existing cutting-edge methods for generating sarcasm explanations for sarcastic multimodal social media posts, certain limitations still exist in \model\ which we address in this section. 

First, we extract external knowledge concepts in a deterministic way, that is, for a given token, it will always be linked to the same external knowledge concept regardless of the context surrounding the token. As shown in Appendix \ref{sec:appendix-error-analysis}, this can lead to the extraction of knowledge concepts that, despite being semantically related to the feature tokens, are not relevant in the given context. We believe that a method devised to dynamically extract relevant knowledge concepts based on the context surrounding a sample can help resolve this and can lead to better sarcasm reasoning. 

Second, we incorporate the target of sarcasm in our model by simply concatenating it with the textual tokens (\S \ref{sec:incorporation-of-target}). While the experimental results prove that incorporating the target of sarcasm in this manner leads to better explanation generations when compared to variants that do not do so, we hypothesize that incorporating this annotation in a more specialized manner can help the model learn more relevant semantic insights. For instance, perhaps fusing it with unimodal or multimodal feature representations in a manner that amplifies the more salient features with respect to the target of sarcasm can lead to a more explicit revelation of the underlying semantic incongruity.

Lastly, we recognize that using an annotated target of sarcasm leads to an additional manually provided input. This constrains the dataset since extending the dataset would now require extra manual annotations for the target of sarcasm as well. However, we suggest that this can be tackled by training another model to learn to generate the target of sarcasm given a multimodal social media post using the current dataset. We can use the output of this model as the input target of sarcasm in \model. This allows for the creation of a generative pipeline that does not require the target of sarcasm to be furnished manually as an input and reduces the potential manual work to be done in case of any extensions to the existing dataset since annotators will not be bound to annotate the target of sarcasm for the newly added samples. While such a pipeline would be useful, it would introduce a new challenge - ensuring that the generated target of sarcasm is up to par since in case it is not, it could end up pointing the explanation generation model in the wrong direction, leading to poor performance.
\section{Ethical Considerations} \label{sec:ethical}
In a task such as this, which revolves around explaining the implicit meaning behind an inherently incongruent form of communication, we must be mindful of certain ethical implications. 

It is important to prevent misinterpretations stemming from inaccurate explanations. Considering that sarcasm is usually used to express one's views in an insulting or mocking way, such misinterpretations can lead to excessive harm by highlighting the mockery used in the utterance more than its actual implicit meaning (which can lead to a feeling of alienation or ridicule among the people being mocked) or even in the form of a complete mischaracterization of what the author meant to express. 

Additionally, we must be cognizant of the fact that sarcasm can be interpreted in different ways in the presence of different cultural contexts. For instance, a simple gesture such as a ``thumbs up'' might signify a job well done in one culture. However, it is completely possible that in another culture, the same gesture is seen as one of disrespect. As a result, any sarcastic utterances involving the use of such a gesture will mean different things across the two cultures. Thus, in order to prevent misinterpretations, it is important to take this cultural aspect of sarcasm into account as well.

\section*{Acknowledgement}
The authors acknowledge the partial support of Infosys Foundation through Center of AI (CAI) at IIIT Delhi.
\bibliography{anthology, custom}

\begin{thebibliography}{33}
\providecommand{\natexlab}[1]{#1}

\bibitem[{Babanejad et~al.(2020)Babanejad, Davoudi, An, and Papagelis}]{babanejad-etal-2020-affective}
Nastaran Babanejad, Heidar Davoudi, Aijun An, and Manos Papagelis. 2020.
\newblock \href {https://doi.org/10.18653/v1/2020.coling-main.20} {Affective and contextual embedding for sarcasm detection}.
\newblock In \emph{Proceedings of the 28th International Conference on Computational Linguistics}, pages 225--243, Barcelona, Spain (Online). International Committee on Computational Linguistics.

\bibitem[{Bedi et~al.(2023)Bedi, Kumar, Akhtar, and Chakraborty}]{bedi-etal-code-mixed-2023}
M.~Bedi, S.~Kumar, M.~Akhtar, and T.~Chakraborty. 2023.
\newblock \href {https://doi.org/10.1109/TAFFC.2021.3083522} {Multi-modal sarcasm detection and humor classification in code-mixed conversations}.
\newblock \emph{IEEE Transactions on Affective Computing}, 14(02):1363--1375.

\bibitem[{Bird et~al.(2009)Bird, Loper, and Klein}]{nltk}
Steven Bird, Edward Loper, and Ewan Klein. 2009.
\newblock \emph{Natural Language Processing with Python}.
\newblock O'Reilly Media Inc.

\bibitem[{Bouazizi and Otsuki~Ohtsuki(2016)}]{Bouazizi-etal-sarcasm-detection-2016}
Mondher Bouazizi and Tomoaki Otsuki~Ohtsuki. 2016.
\newblock \href {https://doi.org/10.1109/ACCESS.2016.2594194} {A pattern-based approach for sarcasm detection on twitter}.
\newblock \emph{IEEE Access}, 4:5477--5488.

\bibitem[{Castro et~al.(2019)Castro, Hazarika, P{\'e}rez-Rosas, Zimmermann, Mihalcea, and Poria}]{castro-etal-2019-towards}
Santiago Castro, Devamanyu Hazarika, Ver{\'o}nica P{\'e}rez-Rosas, Roger Zimmermann, Rada Mihalcea, and Soujanya Poria. 2019.
\newblock \href {https://doi.org/10.18653/v1/P19-1455} {Towards multimodal sarcasm detection (an {\_}{O}bviously{\_} perfect paper)}.
\newblock In \emph{Proceedings of the 57th Annual Meeting of the Association for Computational Linguistics}, pages 4619--4629, Florence, Italy. Association for Computational Linguistics.

\bibitem[{Chakrabarty et~al.(2020)Chakrabarty, Ghosh, Muresan, and Peng}]{chakrabarty-etal-2020-r}
Tuhin Chakrabarty, Debanjan Ghosh, Smaranda Muresan, and Nanyun Peng. 2020.
\newblock \href {https://doi.org/10.18653/v1/2020.acl-main.711} {{R}{\^{}}3: Reverse, retrieve, and rank for sarcasm generation with commonsense knowledge}.
\newblock In \emph{Proceedings of the 58th Annual Meeting of the Association for Computational Linguistics}, pages 7976--7986, Online. Association for Computational Linguistics.

\bibitem[{Desai et~al.(2022)Desai, Chakraborty, and Akhtar}]{Desai_Chakraborty_Akhtar_2022}
Poorav Desai, Tanmoy Chakraborty, and Md~Shad Akhtar. 2022.
\newblock \href {https://doi.org/10.1609/aaai.v36i10.21300} {Nice perfume. how long did you marinate in it? multimodal sarcasm explanation}.
\newblock \emph{Proceedings of the AAAI Conference on Artificial Intelligence}, 36(10):10563--10571.

\bibitem[{Dosovitskiy et~al.(2021)Dosovitskiy, Beyer, Kolesnikov, Weissenborn, Zhai, Unterthiner, Dehghani, Minderer, Heigold, Gelly, Uszkoreit, and Houlsby}]{dosovitskiy2021image}
Alexey Dosovitskiy, Lucas Beyer, Alexander Kolesnikov, Dirk Weissenborn, Xiaohua Zhai, Thomas Unterthiner, Mostafa Dehghani, Matthias Minderer, Georg Heigold, Sylvain Gelly, Jakob Uszkoreit, and Neil Houlsby. 2021.
\newblock \href {https://arxiv.org/abs/2010.11929} {An image is worth 16x16 words: Transformers for image recognition at scale}.
\newblock \emph{Preprint}, arXiv:2010.11929.

\bibitem[{Dubey et~al.(2019)Dubey, Joshi, and Bhattacharyya}]{dubey-2019-sarcasm-generation}
Abhijeet Dubey, Aditya Joshi, and Pushpak Bhattacharyya. 2019.
\newblock \href {https://doi.org/10.1145/3297001.3297043} {Deep models for converting sarcastic utterances into their non sarcastic interpretation}.
\newblock In \emph{Proceedings of the ACM India Joint International Conference on Data Science and Management of Data}, CODS-COMAD '19, page 289–292, New York, NY, USA. Association for Computing Machinery.

\bibitem[{Dubey et~al.(2024)}]{dubey2024llama3herdmodels}
Abhimanyu Dubey et~al. 2024.
\newblock \href {https://arxiv.org/abs/2407.21783} {The llama 3 herd of models}.
\newblock \emph{Preprint}, arXiv:2407.21783.

\bibitem[{Felbo et~al.(2017)Felbo, Mislove, S{\o}gaard, Rahwan, and Lehmann}]{felbo-etal-2017-using}
Bjarke Felbo, Alan Mislove, Anders S{\o}gaard, Iyad Rahwan, and Sune Lehmann. 2017.
\newblock \href {https://doi.org/10.18653/v1/D17-1169} {Using millions of emoji occurrences to learn any-domain representations for detecting sentiment, emotion and sarcasm}.
\newblock In \emph{Proceedings of the 2017 Conference on Empirical Methods in Natural Language Processing}, pages 1615--1625, Copenhagen, Denmark. Association for Computational Linguistics.

\bibitem[{Jiang et~al.(2023)Jiang, Sablayrolles, Mensch, Bamford, Chaplot, de~las Casas, Bressand, Lengyel, Lample, Saulnier, Lavaud, Lachaux, Stock, Scao, Lavril, Wang, Lacroix, and Sayed}]{jiang2023mistral}
Albert~Q. Jiang, Alexandre Sablayrolles, Arthur Mensch, Chris Bamford, Devendra~Singh Chaplot, Diego de~las Casas, Florian Bressand, Gianna Lengyel, Guillaume Lample, Lucile Saulnier, Lélio~Renard Lavaud, Marie-Anne Lachaux, Pierre Stock, Teven~Le Scao, Thibaut Lavril, Thomas Wang, Timothée Lacroix, and William~El Sayed. 2023.
\newblock \href {https://arxiv.org/abs/2310.06825} {Mistral 7b}.
\newblock \emph{Preprint}, arXiv:2310.06825.

\bibitem[{Jiang(2023)}]{jiang-2023-team}
Ye~Jiang. 2023.
\newblock \href {https://doi.org/10.18653/v1/2023.semeval-1.40} {Team {QUST} at {S}em{E}val-2023 task 3: A comprehensive study of monolingual and multilingual approaches for detecting online news genre, framing and persuasion techniques}.
\newblock In \emph{Proceedings of the 17th International Workshop on Semantic Evaluation (SemEval-2023)}, pages 300--306, Toronto, Canada. Association for Computational Linguistics.

\bibitem[{Jing et~al.(2023)Jing, Song, Ouyang, Jia, and Nie}]{jing-etal-2023-multi}
Liqiang Jing, Xuemeng Song, Kun Ouyang, Mengzhao Jia, and Liqiang Nie. 2023.
\newblock \href {https://doi.org/10.18653/v1/2023.acl-long.635} {Multi-source semantic graph-based multimodal sarcasm explanation generation}.
\newblock In \emph{Proceedings of the 61st Annual Meeting of the Association for Computational Linguistics (Volume 1: Long Papers)}, pages 11349--11361, Toronto, Canada. Association for Computational Linguistics.

\bibitem[{Kipf and Welling(2017)}]{kipf2017semisupervised}
Thomas~N. Kipf and Max Welling. 2017.
\newblock \href {https://openreview.net/forum?id=SJU4ayYgl} {Semi-supervised classification with graph convolutional networks}.
\newblock In \emph{International Conference on Learning Representations}.

\bibitem[{Kumar et~al.(2022)Kumar, Kulkarni, Akhtar, and Chakraborty}]{kumar-etal-2022-become}
Shivani Kumar, Atharva Kulkarni, Md~Shad Akhtar, and Tanmoy Chakraborty. 2022.
\newblock \href {https://doi.org/10.18653/v1/2022.acl-long.411} {When did you become so smart, oh wise one?! sarcasm explanation in multi-modal multi-party dialogues}.
\newblock In \emph{Proceedings of the 60th Annual Meeting of the Association for Computational Linguistics (Volume 1: Long Papers)}, pages 5956--5968, Dublin, Ireland. Association for Computational Linguistics.

\bibitem[{Lewis et~al.(2020)Lewis, Liu, Goyal, Ghazvininejad, Mohamed, Levy, Stoyanov, and Zettlemoyer}]{lewis-etal-2020-bart}
Mike Lewis, Yinhan Liu, Naman Goyal, Marjan Ghazvininejad, Abdelrahman Mohamed, Omer Levy, Veselin Stoyanov, and Luke Zettlemoyer. 2020.
\newblock \href {https://doi.org/10.18653/v1/2020.acl-main.703} {{BART}: Denoising sequence-to-sequence pre-training for natural language generation, translation, and comprehension}.
\newblock In \emph{Proceedings of the 58th Annual Meeting of the Association for Computational Linguistics}, pages 7871--7880, Online. Association for Computational Linguistics.

\bibitem[{Li et~al.(2022)Li, Li, Xiong, and Hoi}]{li2022blip}
Junnan Li, Dongxu Li, Caiming Xiong, and Steven Hoi. 2022.
\newblock \href {https://arxiv.org/abs/2201.12086} {Blip: Bootstrapping language-image pre-training for unified vision-language understanding and generation}.
\newblock \emph{Preprint}, arXiv:2201.12086.

\bibitem[{Liu et~al.(2024)Liu, Li, Li, and Lee}]{liu2024improved}
Haotian Liu, Chunyuan Li, Yuheng Li, and Yong~Jae Lee. 2024.
\newblock \href {https://arxiv.org/abs/2310.03744} {Improved baselines with visual instruction tuning}.
\newblock \emph{Preprint}, arXiv:2310.03744.

\bibitem[{Liu et~al.(2020)Liu, Sun, Yu, Zhang, and Xu}]{liu-etal-2020-multistage}
Nayu Liu, Xian Sun, Hongfeng Yu, Wenkai Zhang, and Guangluan Xu. 2020.
\newblock \href {https://doi.org/10.18653/v1/2020.emnlp-main.144} {Multistage fusion with forget gate for multimodal summarization in open-domain videos}.
\newblock In \emph{Proceedings of the 2020 Conference on Empirical Methods in Natural Language Processing (EMNLP)}, pages 1834--1845, Online. Association for Computational Linguistics.

\bibitem[{Loshchilov and Hutter(2017)}]{loshchilov-etal-adamw}
Ilya Loshchilov and Frank Hutter. 2017.
\newblock Fixing weight decay regularization in adam.

\bibitem[{Ma et~al.(2015)Ma, Lu, Shang, and Li}]{ma-etal-cnn-2015}
Lin Ma, Zhengdong Lu, Lifeng Shang, and Hang Li. 2015.
\newblock \href {https://doi.org/10.1109/ICCV.2015.301} {Multimodal convolutional neural networks for matching image and sentence}.
\newblock In \emph{2015 IEEE International Conference on Computer Vision (ICCV)}, pages 2623--2631.

\bibitem[{Mishra et~al.(2019)Mishra, Tater, and Sankaranarayanan}]{mishra-etal-2019-modular}
Abhijit Mishra, Tarun Tater, and Karthik Sankaranarayanan. 2019.
\newblock \href {https://doi.org/10.18653/v1/D19-1636} {A modular architecture for unsupervised sarcasm generation}.
\newblock In \emph{Proceedings of the 2019 Conference on Empirical Methods in Natural Language Processing and the 9th International Joint Conference on Natural Language Processing (EMNLP-IJCNLP)}, pages 6144--6154, Hong Kong, China. Association for Computational Linguistics.

\bibitem[{OpenAI et~al.(2024)OpenAI, Achiam et~al.}]{openai2024gpt4technicalreport}
OpenAI, Josh Achiam, et~al. 2024.
\newblock \href {https://arxiv.org/abs/2303.08774} {Gpt-4 technical report}.
\newblock \emph{Preprint}, arXiv:2303.08774.

\bibitem[{Peled and Reichart(2017)}]{peled-reichart-2017-sarcasm}
Lotem Peled and Roi Reichart. 2017.
\newblock \href {https://doi.org/10.18653/v1/P17-1155} {Sarcasm {SIGN}: Interpreting sarcasm with sentiment based monolingual machine translation}.
\newblock In \emph{Proceedings of the 55th Annual Meeting of the Association for Computational Linguistics (Volume 1: Long Papers)}, pages 1690--1700, Vancouver, Canada. Association for Computational Linguistics.

\bibitem[{Qiao et~al.(2023)Qiao, Jing, Song, Chen, Zhu, and Nie}]{Qiao_Jing_Song_Chen_Zhu_Nie_2023}
Yang Qiao, Liqiang Jing, Xuemeng Song, Xiaolin Chen, Lei Zhu, and Liqiang Nie. 2023.
\newblock \href {https://doi.org/10.1609/aaai.v37i8.26138} {Mutual-enhanced incongruity learning network for multi-modal sarcasm detection}.
\newblock \emph{Proceedings of the AAAI Conference on Artificial Intelligence}, 37(8):9507--9515.

\bibitem[{Schifanella et~al.(2016)Schifanella, de~Juan, Tetreault, and Cao}]{schifanella-etal-2016}
Rossano Schifanella, Paloma de~Juan, Joel Tetreault, and LiangLiang Cao. 2016.
\newblock \href {https://doi.org/10.1145/2964284.2964321} {Detecting sarcasm in multimodal social platforms}.
\newblock In \emph{Proceedings of the 24th ACM International Conference on Multimedia}, MM '16, page 1136–1145, New York, NY, USA. Association for Computing Machinery.

\bibitem[{See et~al.(2017)See, Liu, and Manning}]{see-etal-2017-get}
Abigail See, Peter~J. Liu, and Christopher~D. Manning. 2017.
\newblock \href {https://doi.org/10.18653/v1/P17-1099} {Get to the point: Summarization with pointer-generator networks}.
\newblock In \emph{Proceedings of the 55th Annual Meeting of the Association for Computational Linguistics (Volume 1: Long Papers)}, pages 1073--1083, Vancouver, Canada. Association for Computational Linguistics.

\bibitem[{Speer et~al.(2017)Speer, Chin, and Havasi}]{speer2017conceptnet}
Robyn Speer, Joshua Chin, and Catherine Havasi. 2017.
\newblock \href {http://aaai.org/ocs/index.php/AAAI/AAAI17/paper/view/14972} {Conceptnet 5.5: An open multilingual graph of general knowledge}.

\bibitem[{Tay et~al.(2018)Tay, Luu, Hui, and Su}]{tay-etal-2018-reasoning}
Yi~Tay, Anh~Tuan Luu, Siu~Cheung Hui, and Jian Su. 2018.
\newblock \href {https://doi.org/10.18653/v1/P18-1093} {Reasoning with sarcasm by reading in-between}.
\newblock In \emph{Proceedings of the 56th Annual Meeting of the Association for Computational Linguistics (Volume 1: Long Papers)}, pages 1010--1020, Melbourne, Australia. Association for Computational Linguistics.

\bibitem[{Vaswani et~al.(2017)Vaswani, Shazeer, Parmar, Uszkoreit, Jones, Gomez, Kaiser, and Polosukhin}]{attention-is-all-you-need}
Ashish Vaswani, Noam Shazeer, Niki Parmar, Jakob Uszkoreit, Llion Jones, Aidan~N. Gomez, \L{}ukasz Kaiser, and Illia Polosukhin. 2017.
\newblock Attention is all you need.
\newblock In \emph{Proceedings of the 31st International Conference on Neural Information Processing Systems}, NIPS'17, page 6000–6010, Red Hook, NY, USA. Curran Associates Inc.

\bibitem[{Wang et~al.(2024)Wang, Yeh, and Liao}]{wang2024yolov9}
Chien-Yao Wang, I-Hau Yeh, and Hong-Yuan~Mark Liao. 2024.
\newblock \href {https://arxiv.org/abs/2402.13616} {Yolov9: Learning what you want to learn using programmable gradient information}.
\newblock \emph{Preprint}, arXiv:2402.13616.

\bibitem[{Yao and Wan(2020)}]{yao-wan-2020-multimodal}
Shaowei Yao and Xiaojun Wan. 2020.
\newblock \href {https://doi.org/10.18653/v1/2020.acl-main.400} {Multimodal transformer for multimodal machine translation}.
\newblock In \emph{Proceedings of the 58th Annual Meeting of the Association for Computational Linguistics}, pages 4346--4350, Online. Association for Computational Linguistics.

\end{thebibliography}

\appendix
\section*{Appendix}
\section{Error Analysis}
\label{sec:appendix-error-analysis}

Despite its excellent performance, we observe that \model\ can exhibit different types of errors, which can lead to inaccurate explanations. We lay them down as follows.

\paragraph{Extraction of Irrelevant External Knowledge Concepts:} We recognize that in some cases, the external knowledge concepts extracted by \model\ may not be relevant to the underlying incongruity. This can cause the model to focus on the wrong things when trying to decipher the underlying meaning of sarcasm, leading to inaccurate explanations.

\begin{figure}[!h]
\centering
  \includegraphics[width=0.8\columnwidth]{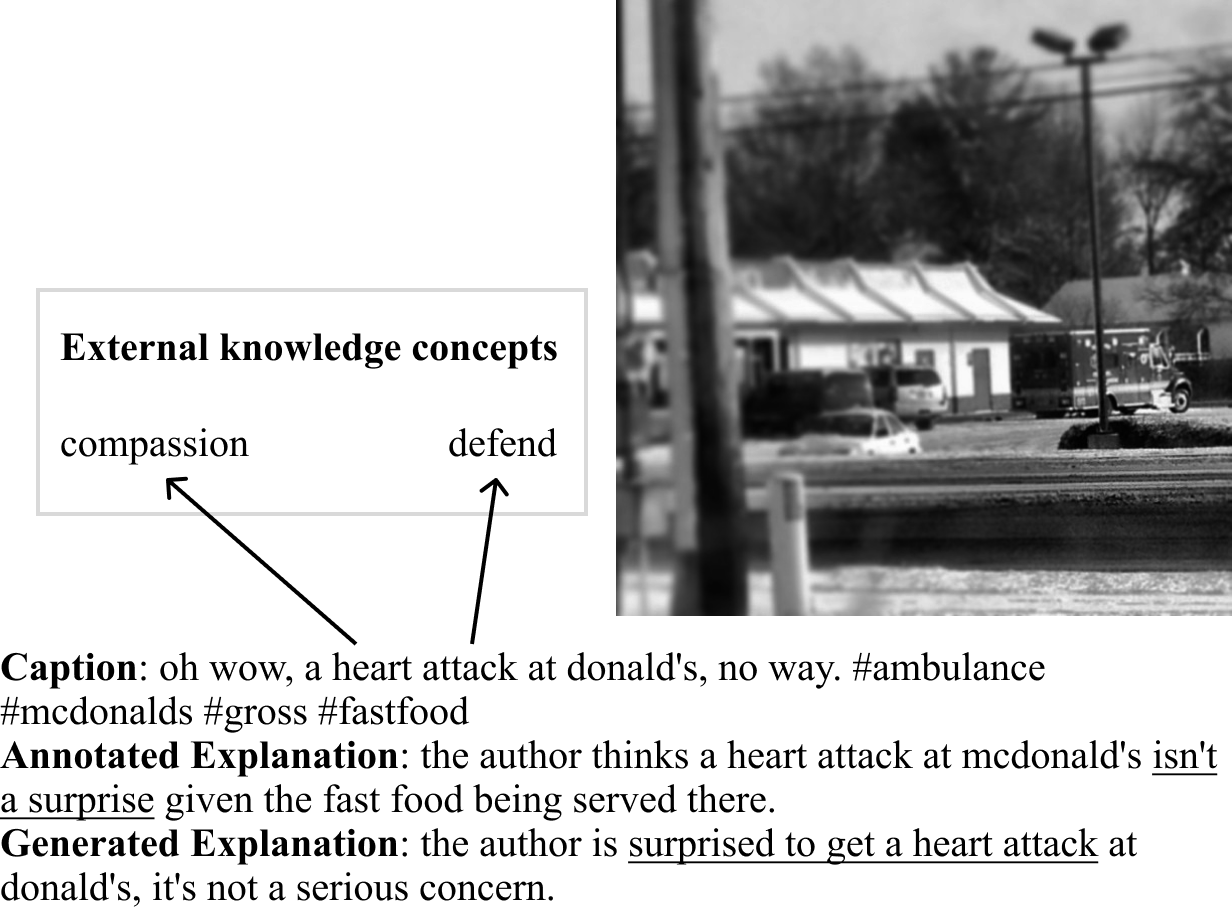}
  \caption{\textbf{Extraction of Irrelevant External Knowledge Concepts:} A sample where \model\ extracted irrelevant external knowledge concepts for the entities vital to understanding the sarcastic incongruity.
}
\label{fig:error-analysis}
\end{figure}

For example, Figure~\ref{fig:error-analysis} shows one such case. The author uses sarcasm to remark that the fast food served at McDonald's is unhealthy and that it would not be a surprise if someone got a heart attack by eating it. We observe that the generated explanation completely misses the sarcasm. It seems to suggest that the author had a heart attack at McDonald's and that it is not a serious concern. 

Upon extracting the external knowledge concepts for this sample from the model, we see that the knowledge concept identified for the word ``heart''  is ``compassion'' and that for ``attack'' is ``defend''. The knowledge concepts make intuitive sense -- a heart is often used to symbolize compassion and the words ``attack'' and ``defend'' are antonyms of each other. However, these concepts are not relevant in this particular context. Instead, the phrase ``heart attack'' should be linked with a knowledge concept such as ``emergency'' since that is much more semantically relevant for this sample. 

We hypothesize that such errors can be mitigated by developing a mechanism that extracts external knowledge concepts by taking the underlying context into account.

\paragraph{Missing External Knowledge Concepts:} We observe that for some samples, the model is unable to extract any external knowledge concepts for entities which are vital to the sarcastic incongruity. This can easily lead to the model not being able to resolve the sarcastic incongruity, especially if it is something that requires world knowledge to understand. 

\begin{figure}[!h]
\centering
  \includegraphics[width=\linewidth]{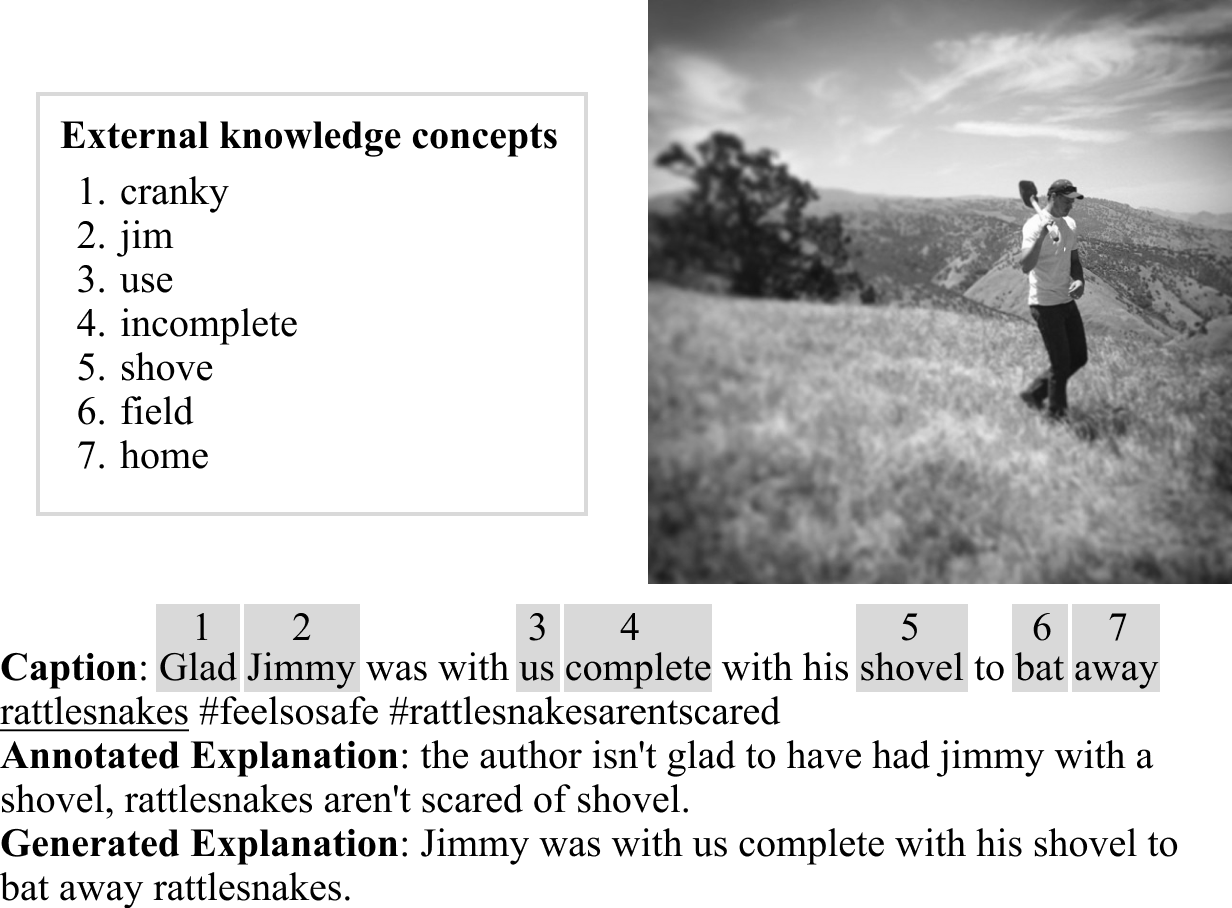}
  \caption{\textbf{Missing External Knowledge Concepts:} A sample where \model\ did not extract any knowledge concepts for the entity relevant to sarcasm. 
}
\label{fig:error-analysis-no-knowledge}
\end{figure}

For example, in Figure~\ref{fig:error-analysis-no-knowledge}, the author of the post is sarcastically mocking ``Jimmy'' for bringing a shovel to scare away any rattlesnakes in the grass. In order to understand why the author is being sarcastic, one must be aware of the fact that rattlesnakes are extremely dangerous creatures and using a shovel to deal with such an animal is not safe. The only way our model would be able to know this information is in the form of external knowledge concepts. In the figure, we have provided the knowledge concepts extracted by the model. Clearly, the model was unable to extract any concepts for ``rattlesnakes'' (underlined in the figure). Due to this, \model\ was unaware of how dangerous rattlesnakes are and how ineffective a shovel is against them. Consequently, it was unable to understand the underlying sarcasm in the post and generated an explanation that simply repeated what the author stated in the caption.

\paragraph{Insufficient OCR Features:} A notable source of error in our model is that it does not explicitly extract features from the textual entities present in the visual modality, aka. OCR text. This is not an issue for the samples which contain a minimal amount of OCR text. However, in samples which rely heavily on it, we realize that \model\ often extracts insufficient features from the OCR text due to which it may generate partially or completely incorrect explanations.

\begin{figure}[!h]
\centering
  \includegraphics[width=\linewidth]{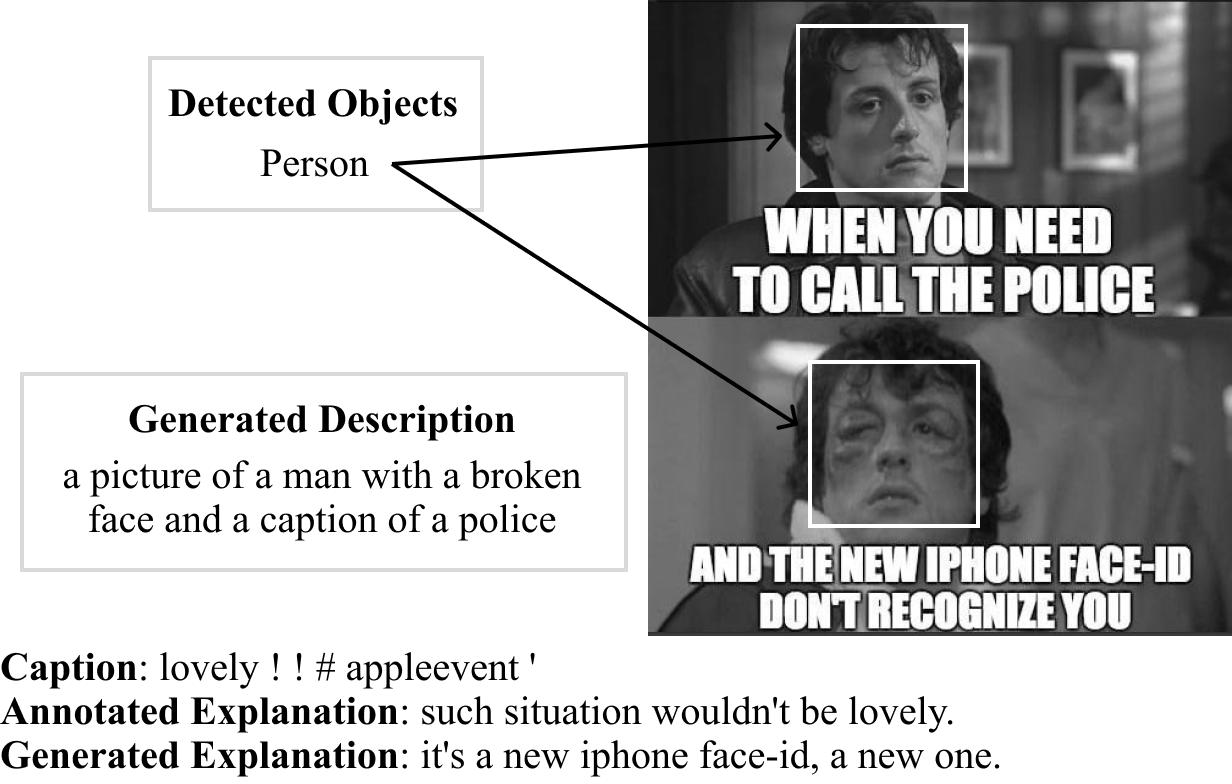}
  \caption{\textbf{Insufficient OCR Features}: A sample containing OCR text where \model\ generated a completely wrong sarcasm explanation. 
}
\label{fig:error-analysis-OCR}
\end{figure}

For instance, Figure~\ref{fig:error-analysis-OCR} depicts a sample where the image contains text and sarcastically ridicules the new ``Face-ID'' technology introduced by Apple in their iPhone. It does so by suggesting the failure of ``Face-ID'' at a time of emergency, leading to some harm to the user. While the explanation generated by \model\ demonstrates that the model has recognized that the post is related to the ``Face-ID'' technology, it is inadequate at explaining the sarcasm behind the post. We observe that the image description generated for this sample refers to the word ``police'' in the image. However, this is not sufficient or relevant enough to allow the model to identify the sarcastic incongruity. 
So, even though \model\ is able to identify OCR features to some extent, there is a definite scope in improving the sufficiency of these extracted features for more apt explanations. 

\begin{figure}[!h]
\centering
  \includegraphics[width=\columnwidth]{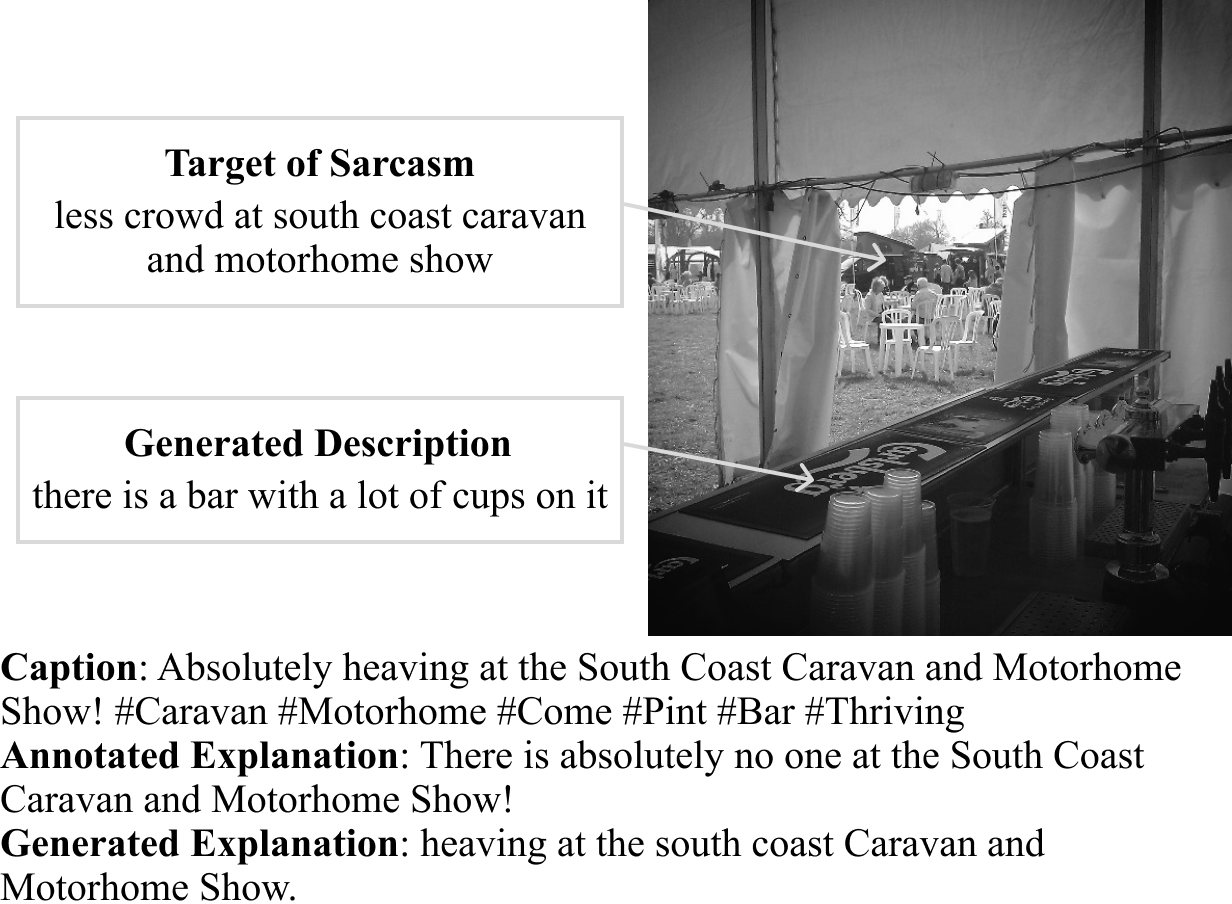}
  \caption{\textbf{Irrelevant Image Description}: A sample where \model\ generated a completely wrong sarcasm explanation due to irrelevant image description. 
}
\label{fig:error-analysis-irrelevant-desc}
\end{figure}
\paragraph{Irrelevant Image Description:} We extract low-level detail from the images in the form of image descriptions generated using BLIP. While this model gives high quality and precise results, it may generate a description that is not quite relevant in understanding the underlying sarcasm. 

For instance, in Figure~\ref{fig:error-analysis-irrelevant-desc}, the author of the post sarcastically makes fun of the scarce crowd at a gathering. This is evident by the empty chairs and tables in the background of the corresponding image. However, the image description used by our model describes the bar in the foreground. As a result, even though the description is objectively accurate, it does not contribute to the semantic understanding of sarcasm in this instance. This, in turn, leads to \model\ generating an incorrect explanation that states that there is ``heaving'' (a big crowd) at the mentioned show when the exact opposite is true instead.

\end{document}